# A Novel Method to Estimate Tilt Angle of a Body using a Pendulum

Anandhu Suresh and Dr. Karnam Venkata Manga Raju

*Abstract*— Most of the advanced control systems use sensor-based feedback for robust control. Tilt angle estimation is key feedback for many robotics and mechatronics applications in order to stabilize a system. Tilt angle cannot be directly measured when the system in consideration is not attached to a stationary frame. it is usually estimated through indirect measurements in such systems. The precision of this estimation depends on the measurements; hence it can get expensive and complicated as the precision requirement increases. This research is aimed at developing a novel and economic method to estimate tilt angle with a relatively less sophisticated and complicated system, while maintaining precision in estimating tilt angle. The method is developed to explore a pendulum as an inertial measurement sensor and estimates tilt angle based on dynamics of pendulum and parameter estimation models. Further, algorithms are developed with varying order of complexity and accuracy to have customization for different applications. Furthermore, this study will validate the developed algorithms by experimental testing. This method focuses on developing algorithms to reduce the input measurement error in the Kalman filter.

## I. INTRODUCTION

We are currently in the era of advancements in field of mechatronics and robotics. Most of the control algorithms in advanced robotics are closed loop system. The robustness of the control algorithm largely depends on the quality of feedback. We are exploring technologies like autonomous vehicles, Intelligent Robots, Intelligent automobiles and much more. Mechatronics, Robotics and Automation has an important role of freeing people from repetitive tasks and enabling them to spend time on creative ventures, which will propel humanity forward. Tilt angle is important feedback for many mechatronic systems like legged robots, automobiles, drones etc... Also, in these types of systems one cannot obtain a tilt angle using direct methods due to the fact that these are not fixed to a stationary frame. So, we have to employ and rely on inertial measurements and sensor fusion algorithms. Usually Inertial Measurement Unit (IMU) measurements are used to estimate tilt angle in these applications. IMU is a sensor cluster measuring accelerations and angular velocities. These measurements have direct one to one correspondence with tilt angle. But it is not an easy task to estimate tilt angles from them.

In 1763,[1] Bayes Thomas came up with a famous theorem in probability, now known as Bayes' theorem [13]. This theorem explained how to mathematically determine the probability change of an event with respect to every new information relevant to that event. This enabled us to update data and minimize error with every new information available. By 1960[2], R. E. Kalman used this theory to develop a filter which became so popular in the science and technology world, The Kalman filter. It enabled us to fuse multiple sensors and predictions to estimate optimal state value. Further, it helped in developing a feasible method to estimate precise tilt angle from an IMU [4]. Many of the technological advances in mechatronics and robotics are made possible by this algorithm. The Kalman filter is an optimal estimation algorithm for linear systems, assuming gaussian noise in the system. In 1997, Simon J. Julier and Jeffrey K. Uhlmann [3] formulated Extended Kalman Filter (EKF), extending Kalman filter to also deal with nonlinear systems. This Enabled to use rotation matrices also as a part of Kalman filter enabling to estimate the complete orientation of a body [5]. Several advanced applications used Kalman Filters [7], [10], [11]. Further development on Kalman filters came as to even include highly nonlinear systems where one cannot approximate the noise is gaussian, The Unscented Kalman Filter.

Furthermore, advancements in MEMS (Micro Electro-Mechanical Systems) technology enabled us to fit a dozen of sensors on a single credit card sized board or even smaller. This made the Inertial measurement systems compact and cheaper. But high precision applications used to keep MEMS sensors away initially as they are susceptible to vibrations. As the estimation algorithms advanced, MEMS sensors are being used in few precision applications. But still the problem persists.

In [4], it is explained how to estimate tilt angle using an IMU by the application of Kalman filters. For decades we used MEMS accelerometers to reference and update the tilt angles. This worked for many applications and also kept the system compact. As the precision required increases or as the environment becomes too noisy, the vibration sensitive nature of MEMS accelerometers pose a disadvantage to the system. This calls for sophisticated and expensive hardware in IMUs, causing the cost to be high. Most of the advanced technologies rely on precise feedback. This increases the cost of the technology causing a resistance to these technologies finding its way to everyday life. The research involved in this paper is aimed at improving the precision of the currently existing systems as well as to provide lesser cost alternatives. This research is primarily aimed at providing a measurement input to Kalman filter which is less susceptible to vibrations

*Research supported by TVS Motor Company.

Anandhu Suresh is with the Chassis Mechatronics Group, TVS Motor Company, Hosur, Tamil Nadu 635109 (phone: +91 9588885155; e-mail: suresh.1@iitj.ac.in; organization e-mail: Anandhu.suresh@tvsmotor.com).

K.V.M Raju is Section Head - Chassis, TVS Motor company, TN 635109 India. (e-mail: Venkata.Raju@tvsmotor.com)

The study begins by investigating about IMU and how the Kalman filter works. A Kalman filter internally combines two random variables with a high variance to provide a random variable with comparatively lower variance than both. The final output variance can reduce with change in the input random variable variance. So, in a Kalman filter, improvement in measurement input will improve the final result too. This study will be introducing a pendulum as an inertial measurement sensor. A pendulum system is susceptible to overshoot and oscillations but very less sensitive to vibrations. This study will explore pendulum dynamics to develop three algorithms to reduce errors of pendulum-based measurement. This study will further validate these algorithms with experiments.

## II. INERTIAL MEASUREMENT UNIT AND KALMAN FILTERS

This study is developed to reduce errors in tilt angle estimates from the Kalman filter while using an IMU. This study emerged from concept of Kalman filters for IMU, hence, for the completeness of this paper this section will provide a brief introduction to IMU (Inertial Measurement Unit) and Kalman filters as referred from [4], [2], [8]. Further, it will also show how the Kalman filter accuracy can be improved using this method.

### A. Inertial Measurement Unit: an introduction

An IMU is a cluster of sensors usually measuring linear accelerations and angular rates. It can be with a varying number of sensors. Most common ones are 6 DOF (Degree of Freedom) and 9 DOF IMUs. A 6 DOF IMU usually has three MEMS accelerometers and three MEMS gyroscopic sensors. A 9 DOF IMU will have an additional three magnetometers to also facilitate precise estimation of yaw angle.

In a three-axis accelerometer, the accelerometer can measure acceleration in three directions orthogonal to each other. Commonly, angle estimation from an accelerometer is based on the fact that the gravity vector changes with respect to the sensor frame when the body is tilted with respect to the earth frame. An Accelerometer is very sensitive to vibrations and thus the angle measured is not very precise.

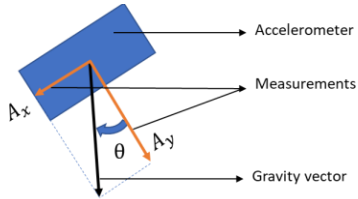

Figure 1.  *Vector diagram for deriving (1)*

$$\theta = \tan^{-1}\left(\frac{A_x}{A_y}\right) \qquad (1)$$

A gyroscopic sensor measures angular velocity. Angular velocity can be integrated to estimate angles. But Integration is prone to drifting error over time. A small bias value in velocity, accumulates over time due to integration making it difficult to estimate roll angle only using gyroscopic sensors.

Sensor fusion methods are used to estimate angles using IMU. Sensor fusion methods mostly rely on combining multiple measurements with different noise signatures to yield a resultant measurement whose error is substantially lesser than all input measurements. One of the commonly used sensor fusion methods in IMU is EKF (Extended Kalman Filter). Kalman filters are great in sensor fusion as it reduces errors in the system by combining prediction and multiple measurements.

### B. Linear Kalman Filters.

Kalman filter is an optimal state estimation algorithm which combines a series of measurement observed over time to produce an estimate measure which can be more accurate than the observed measurements [2].

A linear time invariant system can be written in form:

$$X_k = AX_{k-1} + \omega_k. \qquad (2)$$

$$Y_k = HX_k. \qquad (3)$$

Where $\omega_k$ is zero mean white Gaussian noise which can be assumed to be a normal distribution with variance Q i.e., $\omega_k \sim N(0, Q)$ (A1). $X_k$ is the predicted states at $k^{th}$ instant, similarly $Y_k$ is the predicted measurement at $k^{th}$ instant.

In a system defined by (2) and (3) the covariance changes over time. This can be estimated using (4) as given below.

$$P_k = AP_{k-1}A^T + Q. \qquad (4)$$

Two Gaussian distributions $N(X, \Sigma)$ and $N(X', \Sigma')$, where $(X, X')$ are mean values and $(\Sigma, \Sigma')$ are covariance matrices, can be combined to form:

$$N\left(\frac{X\Sigma' + X'\Sigma}{\Sigma + \Sigma'}, \frac{\Sigma\Sigma'}{\Sigma + \Sigma'}\right). \qquad (5)$$

There are two distributions in above discussed system, predicted measurement i.e., N $(HX_k, HP_kH^T)$ and N$(Z_k, R)$, where $Z_k$ is the actual measured value, measured at $k^{th}$ instant. Using (5) and rearranging, we get:

$$\hat{X}_k = X_k + K(Z_k - HX_k). \qquad (6)$$

$$\hat{P}_k = P_k - KHP_k. \qquad (7)$$

Where $\hat{X}_k$ and $\hat{P}_k$ are our new estimate combining measurements and prediction and K is our Kalman gain given by:

$$K = P_kH(HP_kH + R)^{-1} \qquad (8)$$

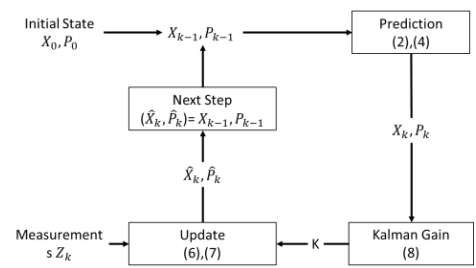

Figure 2.  *Schematic of the Kalman filter Algorithm. The equations corresponding to the process are given in braces*

## C. Kalman Filter for IMU.

For estimating one angle from IMU, a Kalman filter can be designed with states X.

$$X = \begin{bmatrix} \theta \\ b \end{bmatrix} \quad (8)$$

Where $\theta$ and b is angle and the gyroscopic sensor bias respectively. The system equation for IMU is given by:

$$X_k = AX_{k-1} + Bu. \quad (9)$$
$$Y_k = HX_k. \quad (10)$$

$$A = \begin{bmatrix} 1 & -dt \\ 0 & 1 \end{bmatrix}, B = \begin{bmatrix} dt \\ 0 \end{bmatrix}, H = \begin{bmatrix} 1 \\ 0 \end{bmatrix}. \quad (11)$$

Where u is the gyroscopic sensor input and dt is the time step.

## D. Error in Kalman Filter for IMU.

This study is based on the fact that one can improve the error in the Kalman filter output by reducing error in measurement inputs, especially when the error in one of the inputs is too high. The error in output of the Kalman filter is dependent on the error in input measurements. Kalman filters substantially reduce errors in the system as compared to the actual measurements. For one degree of freedom, error after update at $k^{th}$ instant is given by:

$$\sigma_{\hat{X}} = \frac{\sigma_X R}{\sigma_X + R}. \quad (12)$$

In (12), $\sigma_X$ is the predicted measurement error.

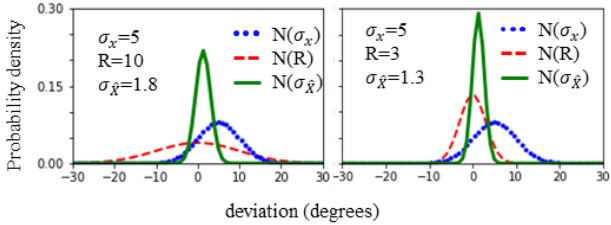

Figure 3. A representative figure to show how the deviation in kalman updated angle improves with reduction in measurement error, assuming a constant prediction error. This figure is aimed at showing how improving the measurement in kalman filters improves final precision.

## E. Extended Kalman Filter

This study can also be extended to an extended Kalman filter. An Extended Kalman filter (EKF) is used when the system in consideration is non-linear. Usually when complete orientation estimation is required, there will be rotational matrices involved in the system, making it non-linear. EKF is similar to nonlinear Kalman filter, where the A and H matrices in (2) and (3) are updated on each iteration as given below

$$A_k = \left.\frac{\partial f_A}{\partial X}\right|_{\hat{X}_{k-1}}. \quad (13)$$

$$H_k = \left.\frac{\partial f_H}{\partial X}\right|_{\hat{X}_{k-1}}. \quad (14)$$

Where $f_A$ and $f_H$ are the nonlinear functions of prediction and measurement respectively.

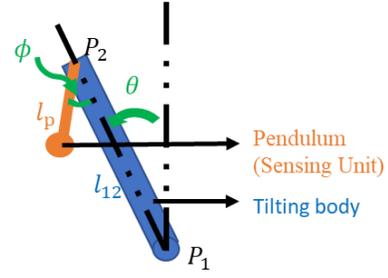

Figure 4. A double pendulum model to derive the dynamics of a pendulum on a tilting body. $l_p$ is the length of pendulum. $l_{12}$ is the length between first pivot(marked as $P_1$) and second pivot(marked as $P_2$). $\theta$ is the tilt angle of body (counter clockwise positive) and $\phi$ is the pendulum angle with respect to body (clockwise positive).

## III. PENDULUM DYNAMICS

This study focuses on using a pendulum as a means of measuring tilt angle of a body. This section explores the dynamics of a pendulum on a tilting body. It can be modelled as a double pendulum of the configuration given in Fig. 4

We can do Euler-Lagrange formulation to derive the equations of the system. A Lagrangian (L) is defined as T-V (Kinetic Energy – Potential Energy). In our system:

$$T = \frac{I_p(-\dot{\phi}+\dot{\theta})^2}{2} + \left(\frac{I_b}{2} + \frac{l_{12}m_p}{2}\right)\dot{\theta}^2. \quad (15)$$

$$V = gl_{bcg}m_b \cos(\theta) + gm_p(l_{12}\cos(\theta) - l_p\cos(\phi-\theta)). \quad (16)$$

Where $I_p$, $I_b$ are the mass moment of inertia and $m_p$, $m_b$ are the masses of pendulum and body respectively. $l_{bcg}$ is the length from $P_1$ to CG (center of gravity) of body.

$$\frac{d}{dt}\left(\frac{dL}{d\dot{q}}\right) - \left(\frac{dL}{dq}\right) + \left(\frac{dD}{d\dot{q}}\right) = 0. \quad (17)$$

Involving Rayleigh's dissipation function in Euler-Lagrange equations, we get (17), where D is the dissipation energy and q is the state variable. One can derive the equations of the given system by solving (17) for different state variables.

$$C\dot{\phi} + I_p(\ddot{\phi} - \ddot{\theta}) + gl_p m_p \sin(\phi - \theta) = 0. \quad (18)$$

## IV. PROPOSED METHODOLOGY

From the previous section, we derived the equations of a double pendulum. We will be measuring the pendulum angle, hence, body tilt ($\theta$) is the unknown variable and pendulum angle ($\phi$) is the known variable. The methodology is development of an algorithm to estimate $\theta$ from $\phi$

### A. Algorithm – 1: only pendulum

It is possible to estimate $\theta$ using only pendulum measurements if we neglect the pseudo force due to angular acceleration of the tilting of the body. Substituting zero for $\ddot{\theta}$ in (18) we get:

$$\theta_{est} = \phi - sin^{-1}\left(\frac{C\dot\phi + I_p\ddot\phi}{gl_pm_p}\right). \quad (20)$$

Where C is the damping coefficient provided by friction at $P_2$ (Fig. 4) and $\theta_{est}$ is the angle estimated using Algorithm-1.

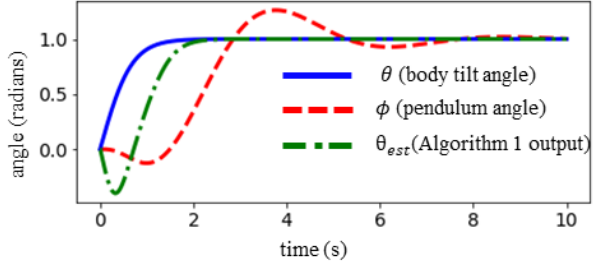

Figure 5. *Simulation result of Algorithm-1 with a sigmoid input. $\theta_{est}$ is computed using (20). This figure also shows the error due to neglection of inertial forces at zero time because of the high $\ddot\theta$ at this moment*

### B. Algorithm-2: pendulum and gyroscopic sensor

This method will involve a gyroscopic sensor along with the pendulum. By differentiating the value from a gyroscopic sensor, we will be able to include pseudo forces due to $\ddot\theta$ also in the system. From (18) we can arrive at the following equation:

$$\theta_{est} = \phi - sin^{-1}\left(\frac{C\dot\phi + I_p(\ddot\phi - G_d)}{gl_pm_p}\right). \quad (21)$$

$$G_d = \frac{(G_k - G_{k-w})}{dt \times w}. \quad (22)$$

In (21) $G_k$ is the gyro value at $k^{th}$ instant and w is the window size considered to avoid noise. dt is the sample time considered. In theory, this estimation exactly coincides with the true value. However, this may not be true in the real-world case due to the delay involved in estimating $\dot\phi, \ddot\phi$ and $G_d$.

### B. Pendulum Parameter Estimation

In Algorithm 1 and 2, it is most accurate if we use measured data to estimate the parameters like $C, I_p$ etc.... We can estimate the parameters of a pendulum if we have true angle data. Parameter estimation can be done through optimization methods. We will be using the popular Newton's method [18] here. Firstly, we have to design a cost function F, which we would want to minimize. The cost function can be the RMS difference between $T_{est}$, and an array of $\theta_{est}(P)$ and T, an array of actual θ

$$F(P) = \sqrt{(T_{est}(P) - T)^2} \quad (23)$$

In (17) P is the parameters involved in case of Algorithm1 and 2 we can take parameters $\frac{C}{m_pgl_p}$ and $\frac{I_p}{m_pgl_p}$.

Newton's method uses multiple iterations to arrive at the optimum value. Iterations are to be continued till the values converge.

$$P_n = P_{n-1} - hF'_{n-1}(P)\left(F''_{n-1}(P)\right)^{-1}. \quad (24)$$

$$F'_n(P) = \frac{1}{\Delta p}\begin{bmatrix}F(P+\Delta P_1) - F(P)\\ \vdots\\ F(P+\Delta P_j) - F(P)\end{bmatrix}\Bigg|_{P_{n-1}}. \quad (25)$$

$$F''_{n-1}(P) = \frac{1}{\Delta p^2}\left[\begin{bmatrix}\Delta f_{11} & \cdots & \Delta f_{1n}\\ \vdots & \ddots & \vdots\\ \Delta f_{n1} & \cdots & \Delta f_{nn}\end{bmatrix}\right]\Bigg|_{P_{n-1}},$$

Where $\Delta f_{jk} = F(P + \Delta P_{jk}) - F(P) \quad (26)$

$$\Delta P_j = \begin{bmatrix}0\\0\\\vdots\\\Delta p(j^{th} element)\\\vdots\\0\end{bmatrix}. \quad (27)$$

$$\Delta P_{jk} = \Delta P_j + \Delta P_k. \quad (28)$$

Equations (24) to (28) shows Newton's method defined for a discrete system where, $\Delta p$ is an infinitesimal value and $\Delta P_j$ is a matrix as shown in (21) with $j^{th}$ value as $\Delta p$ and all other values zero. n is the number of state variables. h is the time step, which one can select.

Once the values converge P can be substituted in (20) or (21) to estimate the body tilt angle.

### C. Pendulum Parameter Estimation for live supervision

For live supervision, we don't have true angle θ, so the cost function cannot be (23). Supervision can be implemented only if we have one more sensor like a gyro. The cost function can be the RMS difference between $\dot T_{est}$, an array of $\dot\theta_{est}(P)$ and G an array of actual gyro values

$$F(P) = \sqrt{\left(\dot T_{est}(P) - G\right)^2}. \quad (29)$$

Equations (24) to (28) is the same for live supervision also.

### D. Algorithm 3 -pendulum and gyro with live supervision

In real life cases there will be delays in system when we $\dot\phi, \ddot\phi$ and $G_d$ as well as when we implement filters. There might also be parameter changes like friction or even sample time. For accommodating these changes, we can monitor the errors and change parameters accordingly using live supervision. Although Algorithm 1 and 2 can also be used along with live supervision, Algorithm 3 is to utilize the complete capability of live supervision techniques. The equation can be modified as below to include error parameters as well as to counter delay.

$$\theta_{est} = \phi - sin^{-1}\left(\frac{K\dot\phi + L\ddot\phi + MG_d}{gl_pm_p}\right) + NG_k. \quad (30)$$

In (30) K, L, M and N are the parameters to be supervised. These parameters will be updated in a regular time interval.

## V. EXPERIMENTATION AND RESULTS

All algorithms are tested using a pivoted tilting body with sensors mounted to it. so that the true value can be measured using an angle potentiometer at the pivot.

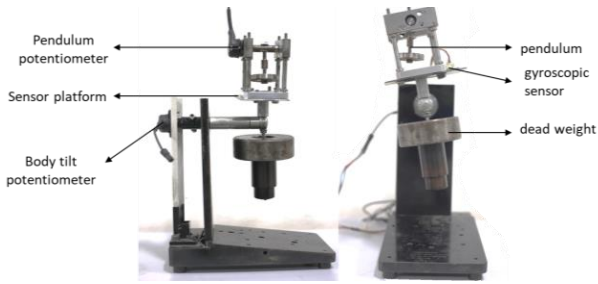

Figure 6.  *A test rig designed to test the accuracy of tilt angle algorithms. The body tilt potentiometer measures the true value so that we can determine the accuracy. The pendulum and gyro constitutes the sensing part. The dead weight is only to keep the sensor platform upright and stable.*

The test rig in Fig. 5 is kept on a motor scooter floorboard with the engine running to mimic an environment with vibration. All the results shown are filtered with a 50$^{th}$ 0rder Low pass FIR filter (20 Hz). Further subsection shows the results with application of algorithms mentioned in the study. Please note that the results are not Kalman filter outputs.

A.  *Pendulum raw angle*

A comparison between true angle, pendulum raw angle and angle derived from accelerometer using (1).

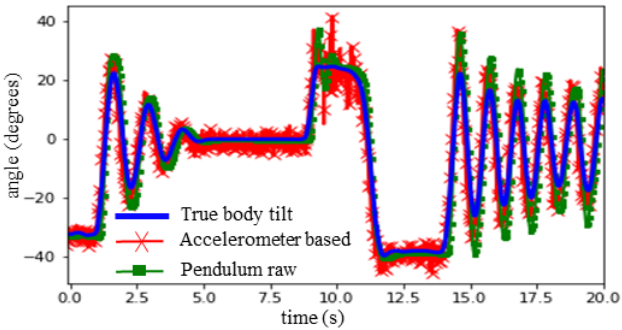

Figure 7.  *Test result showing pendulum raw angle vs angle derived from accelerometer (1). This data shows pendulum performs better in a high vibration environment but it is susceptible to overshoot.*

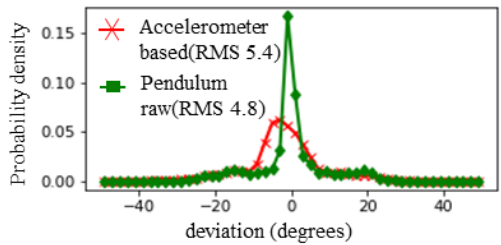

Figure 8.  *Test result showing error in pendulum angle vs angle derived from accelerometer (1). The pendulum raw angle shows lesser error, However, the pendulum error may further increase when inertial forces are high.*

B.  *Algorithm 1: results*

A comparison between true angle and angle derived from algorithm 1(20) and accelerometer (1).

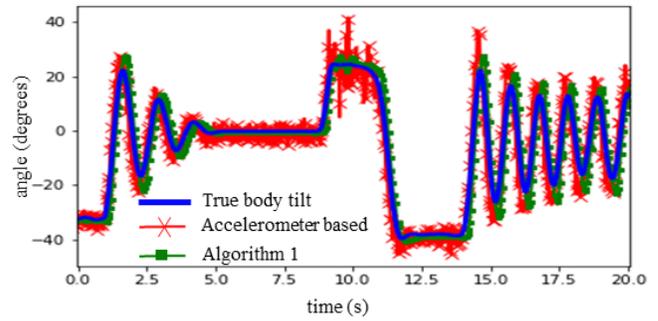

Figure 9.  *Test result showing algorithm 1 implementation using (20). we can still see some overshoot, but it is better than accelerometer input and pendulum raw angle.*

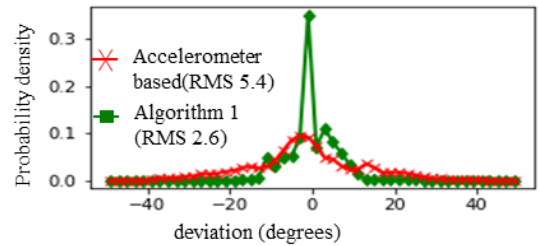

Figure 10.  *Test result showing error in angle estimated using algorithm-1 (20) vs angle derived from accelerometer. Algorithm 1 has substantially lesser noise and error than angle derived from accelerometer*

C.  *Algorithm 2: results*

A comparison between true angle and angle derived from algorithm 2 (21) and accelerometer (1).

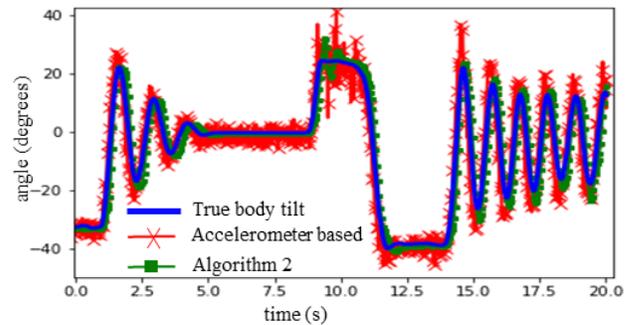

Figure 11.  *Test result showing algorithm 2 implementation using (21). Algorithm 2 results are very close to true angle value*

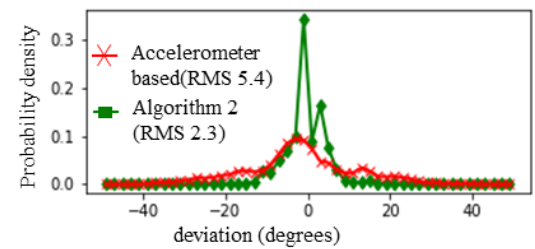

Figure 12.  *Test result showing error in angle estimated using algorithm-2 (21) vs angle derived from accelerometer. This shows a further improvement from Algorithm-1*

*D. Algorithm 3: results*

A comparison between true angle and angle derived from algorithm 2 (21) and accelerometer (1).

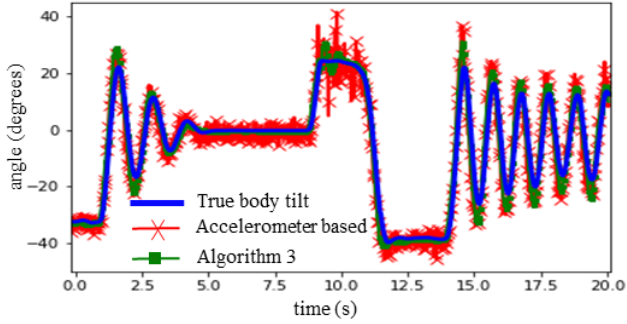

Figure 13. *Test result showing algorithm-3 (30) implementation. This shows a further improvement from Algorithm-2 in measurement delay. This can be observed while comparing with Fig.12*

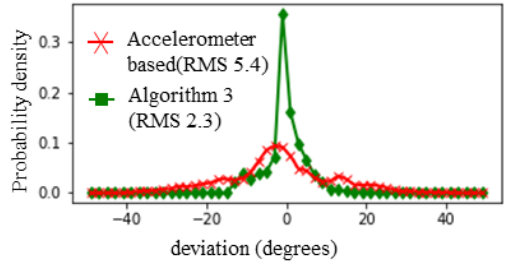

Figure 14. *Test result showing error in angle estimated using algorithm-3 (30) vs angle derived from accelerometer.Eventhough there is an increase in overshoot, This algorith reduces delay. This algorithm can estimate the live parameters of system*

## VI. CONCLUSION

The Algorithms addressed in results algorithm 1, 2 and 3 has shown a capability to reduce certain errors as compared with existing solutions using accelerometers. These can be also used independently for relatively low precision applications. For higher precision applications these can be used as a substitute for accelerometer-based measurement input in Kalman Filters for customized applications or append as an additional measurement to further increase precision of the system. The algorithm complexity is lowest in algorithm-1, further complexity increases in algorithm 2 and 3. The live supervision mentioned in section IV C can contribute to reasonable reduction in delay as well as to account for the parameter change over time. This method can also be used along with algorithm-1 and 2. Algorithm-3 has comparatively less delay but this comes with a tradeoff of overshoot. This can be used for applications requiring less delay in measurements.

Results of the Kalman filter are not shown as the Kalman filter error is dependent on many parameters. From (12) and Fig. 3, we know that a reduction in error in measurement helps improve the final accuracy of updated output.

A simple and relatively unsophisticated system being capable of tilt angle sensing, as shown from results, can help improve the reachability of technology to the public. Also, this explores the possibility of sensing states using systems with known dynamics attached to the body in consideration.

This study is currently in the preliminary stage. Further work on this method will be on extending this methodology to Extended Kalman Filters (EKF). Another area will be reducing the size of the pendulum to make the system further compact as well as to investigate the possibility of a pendulum MEMS and its advantages and disadvantages.

## APPENDIX

The multivariate Gaussian normal distribution probability density function is given below for $X \in \mathbb{R}^k$ AND $\Sigma \in \mathbb{R}^{k \times k}$

$$N(X, \Sigma) = \frac{\Sigma^{-\frac{1}{2}}}{(2\pi)^{\frac{k}{2}}} e^{\frac{-\left((x-X)\Sigma^{-1}(x-X)^T\right)}{2}}. \quad (A1)$$

A sigmoid function $S(x)$ is given by

$$S(x) = \frac{1}{1+e^{-x}}. \quad (A2)$$

## ACKNOWLEDGMENT

Anandhu Suresh acknowledges Dr. Kaushal A Desai and Dr. Suril V Shah, Associate professors, Indian Institute of Technology Jodhpur, for mentoring and motivating me in taking up the field of mechatronics. Further, Anandhu Suresh acknowledges M Barath, Vehicle Dynamics Group, TVS Motor company for all the support and guidance he offered as well as for having belief in the author at the spawn of this research. Anandhu Suresh is indebted to his peers, Abhishek Sharma and M. Santhosh, Assistant Managers, TVS Motor Company for all the support they offered to make this research possible. Anandhu Suresh also extends his gratitude to V. Sai Praveen, Vehicle Dynamics Group, TVS Motor for the support offered in the research methodologies and documentation.